\documentclass[conference]{IEEEtran}
\IEEEoverridecommandlockouts
\usepackage{cite}
\usepackage{amsmath,amssymb,amsfonts}
\usepackage{algorithmic}
\usepackage{graphicx}
\usepackage{textcomp}
\usepackage{xcolor}
\usepackage{booktabs}
\usepackage{makecell}
\usepackage{indentfirst}
\def\BibTeX{{\rm B\kern-.05em{\sc i\kern-.025em b}\kern-.08em
    T\kern-.1667em\lower.7ex\hbox{E}\kern-.125emX}}
\begin{document}

\title{IndGIC: Supervised Action Recognition \\under Low Illumination\\
}

\author{\IEEEauthorblockN{Zeng Jingbo}
\IEEEauthorblockA{\textit{School of Electrical and Electronic Engineering} \\
\textit{Nanyang Technological University}\\
Singapore \\
ZENG0143@e.ntu.edu.sg}
}

\maketitle

\begin{abstract}
Technologies of human action recognition in the dark are gaining more and more attention as huge demand in surveillance, motion control and human-computer interaction. However, because of limitation in image enhancement method and low-lighting video datasets, e.g. labeling cost, existing methods meet some problems. Some video-based approached are effect and efficient in specific datasets but cannot generalize to most cases while others methods using multiple sensors rely heavily to prior knowledge to deal with noisy nature from video stream. In this paper, we proposes action recognition method using deep multi-input network. Furthermore, we proposed a Independent Gamma Intensity Corretion (Ind-GIC) to enhance poor-illumination video, generating one gamma for one frame to increase enhancement performance. To prove our method is effective, there is some evaluation and comparison between our method and existing methods. Experimental results show that our model achieves high accuracy in on ARID dataset.
\end{abstract}


\section{Introduction}
Strong anti-interference action recognition is needed in multiple areas like interaction, motion control and surveillance. This huge demand and vast potential for development prospects have spawned many advanced technologies. Ryoo \emph{et al.}\cite{b1} provided AssembleNet for searching multi-stream neural connectivity for video understanding. Zhang \emph{et al.}\cite{b2} applied a 4D video-level convolutional neural network(V4D) to consider spacial-temporal features. Feichtenhofer C \emph{et al.}\cite{b3} raised X3D, Li \emph{et al.}\cite{b4} provided TEA, Fayyaz M \emph{et al.}\cite{b5} gave a 3D CNN with temporal features resolution. 

However, most approaches consider videos captured under a good lightening condition which have high-quality components and ignore those taken under a low-light condition.With inevitable constraints of technology and environment, this scenario brings some difficulties to human action recognition(HAR). Most existing methods do not have generalization about action recognition from day-time to night-time. We raise a method of preprocessing poor illumination videos and a new structure of neural network combining optical flow, sampling images and original videos.

Technically speaking, data collection modalities can be divided into the following categories, using Rader, point-cloud, RGB image, and skeleton as shown in Fig. 1. RGB image\cite{b6} can provide a wealth of information, and it is generally the most widely used data, as it is easy to obtain and operate. Skeleton\cite{b7} can accurately describe the posture of human body, meanwhile it encode with body joint, providing high-level performance in few-objects scenarios. Point-cloud\cite{b8} which is captured by depth-sensing camera, can provide 3D information like depth and distance, is a key technology in spatial modelling, autonomous driving and robot navigation. Radar\cite{b9} is less commonly used. It can realize through-wall recognition and it is insensitive to interference like noise and change of illumination.\\

\begin{table}[ht]
\centering
\label{tabelsample}
\begin{tabular}[t]{lcc}
\toprule
Modalities&Cons\\
\midrule
RGB image&sensitive to illumination and noise\\
skeletonl&lack of information like color and texture\\
Point cloud&large computation, lack of color and texture\\
Radar&need expensive device\\
\bottomrule
\\
\end{tabular}
\caption{Cons of Modalities of data collect}
\end{table}

In this work we introduce a deep fusion network for robust action recognition through multi-stream inputs. Our work is based on the theory of optical flow and temporal information in videos. There are three input in our framework, original video, enhanced video, and optical flow. Optical flow refers to the movement of the target pixel in the image caused by the movement of the object or the camera in two consecutive frames of the image. The extraction of optical flow in a whole video can indicate a motion trajectory of the key point. Assuming that trajectories of body joints are similar to the same category of motion, a serious of optical flow can be seen as parts of input to our neural network. Spatial information and temporal information can be learnt from video stream. We consider this paper to have the following distributions: 1).We propose a novel framework to fuse deep correlations from multiple stream input in labeled domain, which is useful and effective in human action recognition in the low illumination scenario. Experimental results have shown out method have high accuracy on our dataset. 2).We propose a method of enhancement called Independent-GIC (Ind-GIC). This method provides a good enhancement effect for videos under poor lighting conditions, and the enhancement result is close to the actual environment with sufficient illumination.

\section{Related work}

\subsection{Image Enhancement}
Suboptimal lighting condition caused by technical constraint or environmental reasons will cause incorrect information transmission. This is the first difficulty encountered in the process of dealing with human action recognition in practical application. As it has become the focus of many researchers, large demand has led to many advanced technologies. 

Wei C \emph{et al.}\cite{b10} proposed a end-to-end network Retinex-Net, considering that oberved image can be divided into the reflectance and illumination. One important assumption is that reflectance shared by paired low/normal-light imagesis consistent, and the illumination is smooth. These key constraints can effectively reduce the number of parameters required in the training process, meanwhile the structure becomes simpler. Jiang Y\emph{et al.}\cite{b11} provided first unpaired training network completing low light image enhancement. This removes dependency on paired training samples, hence using data from different domains with different modalities can be realized. Moreover, they introduce a self-regularized attention mechanism contributing to model success. K.Lu\emph{et al.}\cite{b12} gave a two-branch exposure-fusion network called TBEFN. These two branches can enhance slightly and heavily distortion image using a generation-and-fusion strategy.

As both supervised learning and unsupervised learning existing some difficulties like limited generation and unstable training process, zero-shot learning was proposed. The key theory of zero-shot learning is that it donot need paired or unpaired training data, only learning strategy method from testing image, avoiding the risk of overfitting. Guo \emph{et al.}\cite{b13} provided an approach called Zero-DCE. This method abstracts end-to-end image mapping reformulate to a specific curve estimation problem. Mu \emph{et al.}\cite{b14} gave a more effective method derived from Zero-DCE called Zero-DCE++. This approach introduced CSPNET into U-NET, separeting feature map into two parts, then reformulating through cross-phase connection structure, which can realize less computation. 

\subsection{Action Recognition Model}
Recently there are large number of action recognition model shows their advanced performance on different dataset. Gowda W N \emph{et al.}\cite{b15} proposed SMART structure for selecting useful frames in recognition, reaching 98.64$\%$ accuracy on UCF101. Unlike previous work always consider one frame at a time, they consider frames jointly, which leads to a more efficient recognition as useful frames are distributed discretely throughout whole video. Wang L \emph{et al.}\cite{b16} applied DEEP-HAL with ODF+SDF (I3D) to predict object descriptors to realize self-supervising action recognition with 87.560$\%$ accuracy on HMDB-51. Theoretical fundamental is probability distribution function, leading to the capture of statistical moments on descriptors. Most existing researches assuming all the training data are positive samples. Xia B \emph{et al.}\cite{b17} notice the discrimination between positive samples and negative samples using NSNET to enlarge influence 
 
\begin{figure}[h]
\begin{minipage}{0.32\linewidth}
\vspace{3pt}
\centerline{\includegraphics[width=\textwidth]{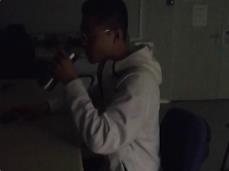}}
\centerline{(a)Origin}
\end{minipage}
\begin{minipage}{0.32\linewidth}
\vspace{3pt}
\centerline{\includegraphics[width=\textwidth]{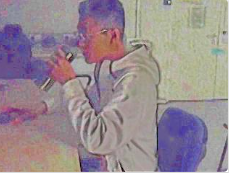}}
\centerline{(b)Retinex-NET}
\end{minipage}
\begin{minipage}{0.32\linewidth}
\vspace{3pt}
\centerline{\includegraphics[width=\textwidth]{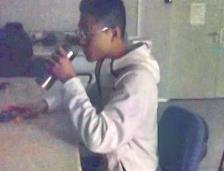}}
\centerline{(c)EnlightenGAN}
\end{minipage}
\end{figure}
\begin{figure}[h]
\begin{minipage}{0.32\linewidth}
\vspace{3pt}
\centerline{\includegraphics[width=\textwidth]{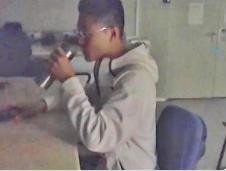}}
\centerline{(d)TBEFN}
\end{minipage}
\begin{minipage}{0.32\linewidth}
\vspace{3pt}
\centerline{\includegraphics[width=\textwidth]{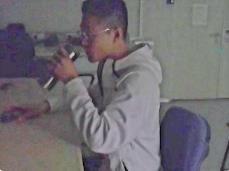}}
\centerline{(e)Zero-DCE}
\end{minipage}
\begin{minipage}{0.32\linewidth}
\vspace{3pt}
\centerline{\includegraphics[width=\textwidth]{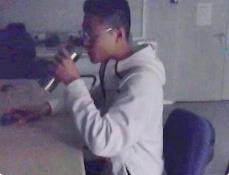}}
\centerline{(f)Zero-DCE++}
\end{minipage}
\caption{Comparisons on a typical poor-illumination image. Using color, naturalness and brightness to judge the performance of different enhancement method. Zero-DCE both provides close and highly reductive result on the using dataset. }
\label{fig}
\end{figure}
\noindent of positive samples in training process, eliminating negative effect produced by negative samples. The accuracy of their approach reaches 94.300$\%$ on ActivityNet. In addition, some other models like VideoMAE\cite{b18}, Uniformer-B\cite{b19}, M$\&M$\cite{b20}, PoseC3D\cite{b21} and DirecFormer\cite{b22} all give advanced results.

\begin{figure}[h]
\begin{minipage}{0.48\linewidth}
\vspace{3pt}
\centerline{\includegraphics[width=\textwidth]{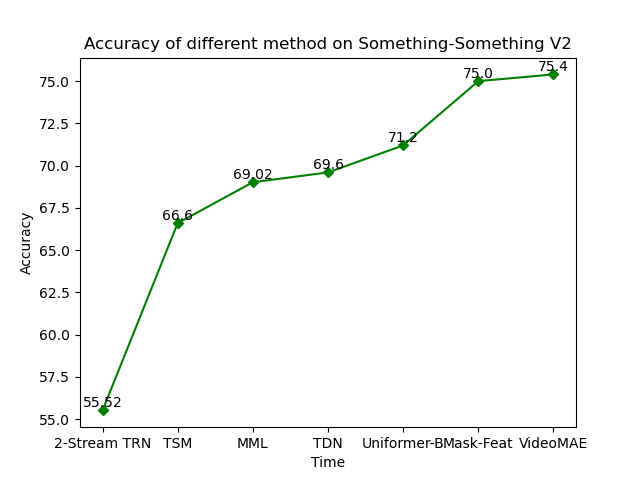}}
\centerline{(a)Something-Something V2}
\end{minipage}
\begin{minipage}{0.48\linewidth}
\vspace{3pt}
\centerline{\includegraphics[width=\textwidth]{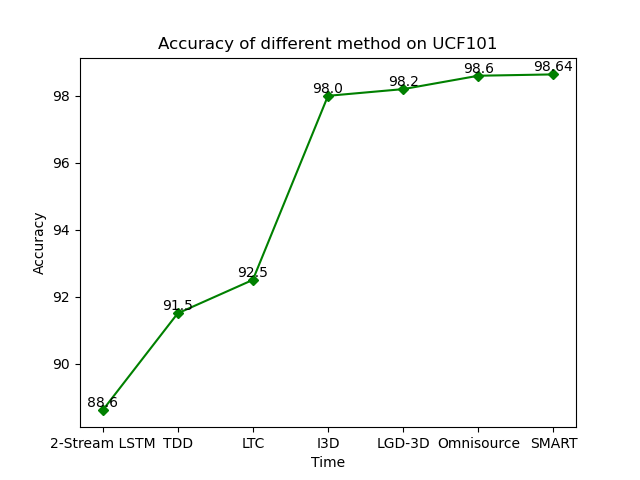}}
\centerline{(b)UCF101}
\end{minipage}
\end{figure}
\begin{figure}[h]
\begin{minipage}{0.48\linewidth}
\vspace{3pt}
\centerline{\includegraphics[width=\textwidth]{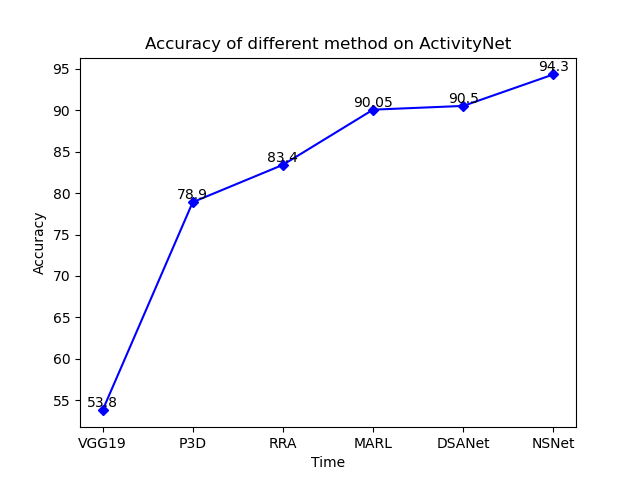}}
\centerline{(c)ActivityNet}
\end{minipage}
\begin{minipage}{0.48\linewidth}
\vspace{3pt}
\centerline{\includegraphics[width=\textwidth]{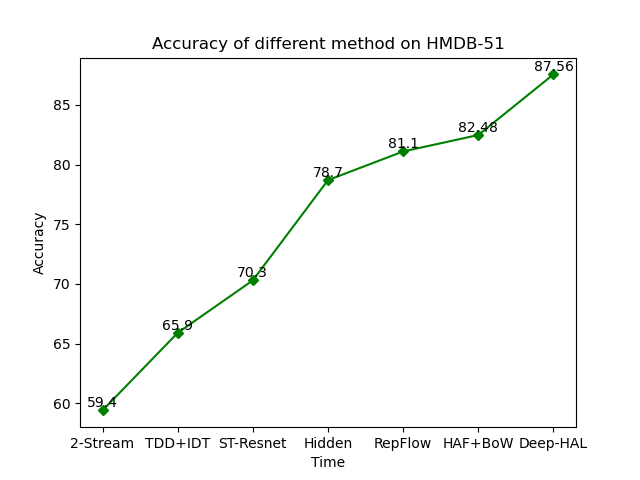}}
\centerline{(d)HMDB-51}
\end{minipage}
\caption{Some existing method testing on different video dataset.}
\label{fig}
\end{figure}

\subsection{Optical Flow}
Optical flow is used to describe motion mode of key points in a scenario with relative motion. Differential method can be obtained from brightness constancy constraint. Components of the motion in X and Y directions can be computed by applying differential computation to constraint function. 

\begin{figure*}
\begin{minipage}{1.03\linewidth}
\vspace{3pt}
\centerline{\includegraphics[width=\textwidth]{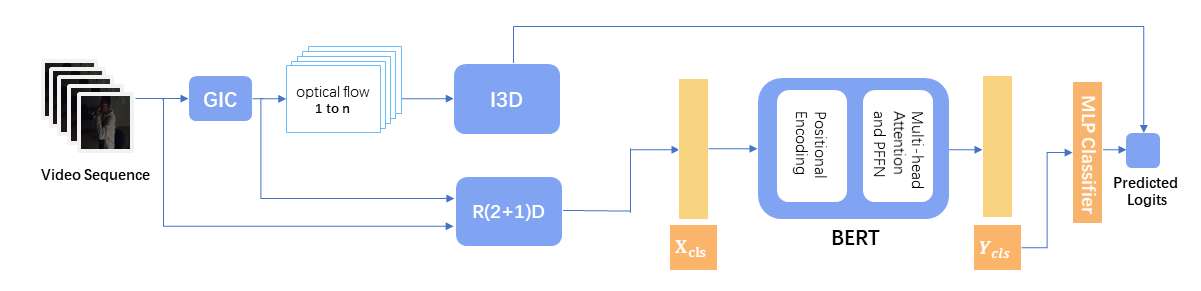}}
\end{minipage}
\caption{The architecture of network. The input is a video sequence with low lighting. The dark input video is enhanced from a new method of Gamma Intense Correction (GIC) designed by ourselves, getting an adaptive gamma value through learning. We design a multi-stream method, an I3D sub-network is used to training optical flow and another R(2+1)D network is used to learn features from both dark and light videos. These two kinds of features are fused in BERT, generating a result which can be used to classify.}
\label{fig}
\end{figure*}

Many optical flow estimation approaches are based on this, \emph{e.g.} Lucas-Kanade and Horn-Schunck\cite{b23}. Difference method to estimate optical flow based on the theory computing difference between two continuous frames on a pixel or a block like Block-Based method\cite{24}. Recent years some researchers have applied neural network into learning optical flow, providing some algorithm with low error rate\cite{25}\cite{26}\cite{27}\cite{28}\cite{29}. 

\subsection{3D CNN + BERT}

3D CNNs are networks derived from 3D convolution throughout whole frame architecture. To satisfy the demand of processing 3D data, the filters are designed in 3D, with different dimensions of channels and temporal information representation. Not as similar as temporal fusion techniques, 3D CNNS take a structured approach in processing temporal information. Before the appearance of 3D CNN, temporal modeling was always completed by extra optical flow or a temporal pooling layer was used directly. However, these methods both have some constraints, limited by 2D convolution, and very difficult to bring temporal information into channel dimension. More usage of 3D CNN does not mean it does not have any downside, \emph{e.g.} huge computational costs and memory demand.

BERT is a pre-trained deep bidirectional transformer for language understanding provided by Devlin J \emph{et al.}\cite{36}. Initially, it is just used to semantic analysis, as it has a unique attention mechanism that can consider input in the time dimension. Google researchers applied BERT to video analysis and provided a model called VideoBERT\cite{37}. They proposed a joint visual-linguistic self-supervised model to learn high-level features, while most existing approaches at that time just provided low-level representations on unlabeled data. 

The first time that 3D CNN combined with BERT is used in the action recognition domain is provided by Kalfaoglu M \emph{et al.}\cite{35}. To realize the fusion of 3D CNN and late temporal modeling, they used a Bidirectional Encoder Representations from Transformers (BERT) layer to replace Temporal Global Average Pooling (TGAP) layer. The advantage of using BERT is that the attention mechanism in BERT can provide a better usage of temporal information. 

\section{Methodology}
\subsection{DarkLight conversion}
Existing image enhancement method like Gamma Intensity Correction (GIC) LIME\cite{30} and deep learning KinD\cite{31} just try to find a most suitable for all the training samples, but this seems can just provide a average performance, ignoring the possibility of highlighting each video to the best effect. Because of this, a simple but really effective conventional method is adopted to give every video or every frame a specific gamma value, derived by Equation 1.
\begin{equation}
Ind-GIC(P) = \frac{\sum p_{max}(\frac{p}{p_{max}})^\frac{1}{\gamma}}{N}
\end{equation}
\noindent where $P$ indicates value of pixel with the range of [0,255], $P_{max}$ is the maximum intensity of input, $\gamma$ means the degree of luminance and $N$ is the number of frames or videos. 

Our method is be regarded as one kind of adaptive GIC. Given the input dark video as a sequence of clips denoted by $I \in R^{N×H×W×3}$, where N is number of frames or videos, H and W are height and width of a frame and 3 indicates RGB channels. First we collect some pictures under normal light and get their $\gamma$. To train our network, we adjust the brightness from $\gamma$ to $\frac{1}{\gamma}$ and set $\gamma$ as their labels respectively. We hope the network can output $\gamma$ after training period. Frames after enhancement denoted by $I_{Ind-GIC} \in R^{N×H×W×3}$. $I$ and $I_{Ind-GIC}$ will fuse together in R(2+1)D subnetwork,

\begin{equation}
Feature_{Dark} = R(I)
\end{equation}

\begin{equation}
Feature_{Light} = R(Ind-GIC)
\end{equation}

\begin{figure}[h]
\centering
\begin{minipage}{0.48\linewidth}
\vspace{3pt}
\centerline{\includegraphics[width=\textwidth]{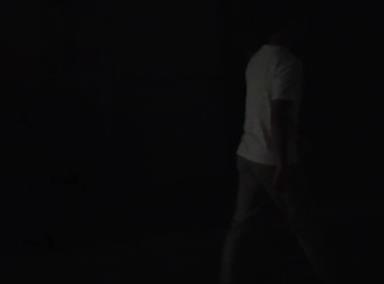}}
\centerline{(a)}
\end{minipage}\begin{minipage}{0.58\linewidth}
\vspace{3pt}
\centerline{\includegraphics[width=\textwidth]{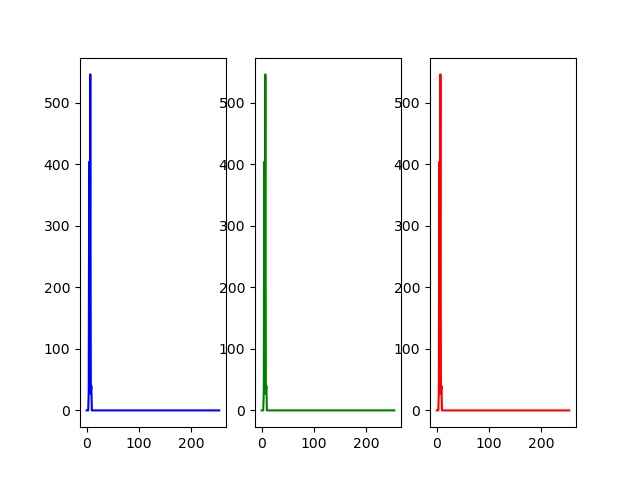}}
\centerline{(b)}
\end{minipage}
\end{figure}
\begin{figure}[h]
\centering
\begin{minipage}{0.48\linewidth}
\vspace{3pt}
\centerline{\includegraphics[width=\textwidth]{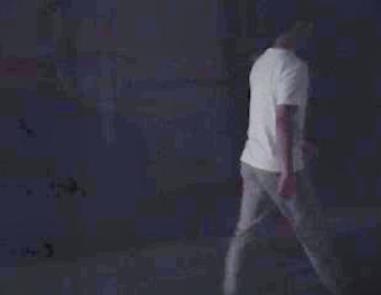}}
\centerline{(c)}
\end{minipage}\begin{minipage}{0.58\linewidth}
\vspace{3pt}
\centerline{\includegraphics[width=\textwidth]{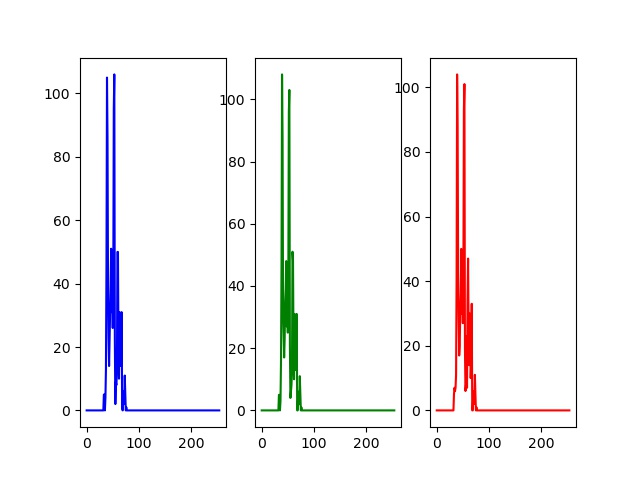}}
\centerline{(d)}
\end{minipage}
\caption{(a) A frame in ARID. (b) Histogram of dark image in RGB channels. (c) The same frame after Ind-GIC. (d) Histogram of enhanced image in RGB channels}
\label{fig}
\end{figure}

\begin{equation}
Feature_{R} = Feature_{Dark} \cap Feature_{Light}
\end{equation}

\noindent where $Feature_{Dark}$ is feature from unenhanced image, $Feature_{Light}$ indicates feature from enhanced image, $Feature_{R}$ means fusion of $Feature_{Dark}$ and $Feature_{Light}$, and $R()$ indicates output of R(2+1)D.

\subsection{Preprocessing}
The vast majority of videos represent action that occurs within only a few specific frames ,which means traditional sampling method with a fixed frequency will cause information loss. Based on this idea, the delta sampling\cite{32} was proposed.  First it calculate base sampling rate $\omega$, then add a $delta$ from a range $[\alpha, \beta]$.   

\begin{equation}
\delta = uniform[\alpha, \beta]
\end{equation}

In most cases $\alpha$ is zero because if it smaller than zero, it indicates cropping the video which may lead to loss on important frames. Then we set $\sigma$a as the largest sampling rate that can be tolerated.

\begin{equation}
S = min[\omega + \delta, \sigma]
\end{equation}

\noindent where $S$ indicates real sampling rate of input video.

Finally, to keep consistent with the number of original output, we adding blank frames before and after valid frame sequence.

\begin{equation}
p_1 = uniform[0, \frac{N}{\omega} - \frac{N}{S} )
\end{equation}

\begin{equation}
p_2 = \frac{N}{\omega} - p_1
\end{equation}

\noindent where $p_1$ and $p_2$ indicates the number of blank before and after valid frames.

Therefore, the whole sequence can be denoted by
\begin{equation}
X = (p_1, \frac{N}{S}, p_2)
\end{equation}

\begin{figure}
\centering
\begin{minipage}{1.0\linewidth}
\vspace{3pt}
\centerline{\includegraphics[width=\textwidth]{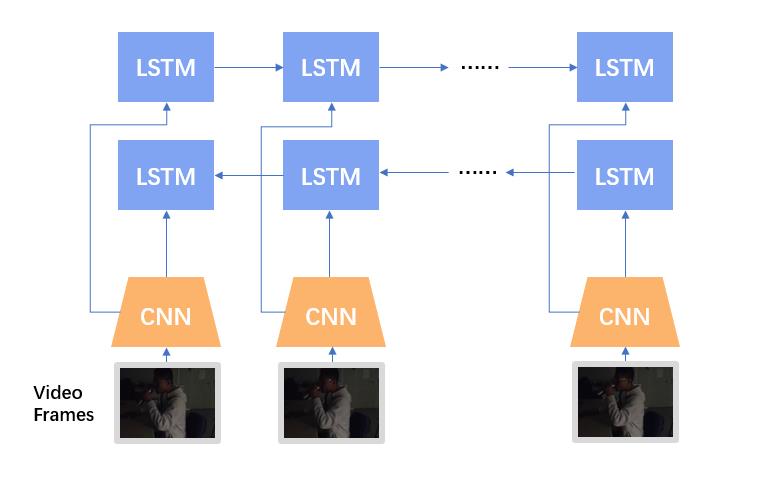}}
\end{minipage}
\caption{Structure of Ind-GIC.}
\label{fig}
\end{figure}

After sampling, we suggest capturing more important information in an image by scaling and cropping can help us get a better training result. The initial size of image is 170*128 pixels. We tried Three methods in doing this. The first one is center cropping, we crop the middle 112*112 pixels in a picture. Next approach we called it maxcenter, using 128*128 pixels in one image. The third one is giving up cropping, resizing the image to 128*128 directly. We found these two practices did have some effect on the accuracy of results, and the corresponding data and analysis will be shown in next section.

\subsection{Network Architecture}
We fed original videos and enhanced videos to R(2+1)D to extract spatial and temporal features, injected optical flow to I3D to learn another feature. These two network are stitched together and their are fused in BERT. The multi-head attention mechanism gives a much better performance than separated features. One advantage of using BERT is that the BERT model can highly preserve temporal information in the whole learning period. After that one linear MLP classification is appended to classify different labels of input videos.

\section{Experiments}
\subsection{Experimental Details}
Our experiment is based on Pytorch. We conduct experiments on a low-lighting dataset ARID\cite{33} which consists of 3784 video clips in 11 action categories. We change the required conditions one by one through the control variables and record Top-1 and Top-5 accuracy in every experiment. For feature extractor, we tests ResNet-18 and I3D with different sampling method like traditional method and Delta sampling. The input of extractor is 16 or 64. Then we test different resizing method like center clipping, maxcenter clipping and scaling. The size of input sequences can be denoted by $X^{3×N×112×112}$ or $X^{3×N×128×128}$, where 3 means three RGB channels, N indicates length of sequences that can be chosen from 16 and 64, 112 and 128 is the number of pixels. Whether there is a $L2_norm$ after pooling is also considered. Next we test the enhancement performance of Ind-GIC on a small dataset containing 20,000 frames sampled from UCF101\cite{34}. Finally we test our whole both-flow network on the testing dataset.

\begin{table}[ht]
\centering
\label{tabelsample}
\begin{tabular}{cccccc}
\toprule
Method & is-dark & Sample & Clip & Top-1 & Top-5\\ 
\midrule
ResNet-18 & False & Orgin & 16 & 47.8$\%$ & -\\
I3D-RGB & True & Origin & 16 & 56.56$\%$ & -\\
I3D-Flow & True & Origin & 16 & 65.31$\%$ & -\\
I3D-Flow & True & Delta & 16 & 65.63$\%$ & -\\
I3D-Flow & True & Delta & 64 & 68.75$\%$ & 1\\
I3D-Flow & True & Origin & 64 & 30$\%$ & 74$\%$\\
I3D-Flow & True & Delta-fixed & 64 & 68.75$\%$ & 99$\%$\\
\bottomrule
\\
\end{tabular}
\caption{Top-1 and Top-5 accuracy.}
\end{table}

\subsection{Results and Comparisons}
First we examine the basic accuracy of testing the framework on the basis of non-optimized models\cite{33} in Table 1. From basic results shown above we found that change sampling method and the length of clipping have distribution to Top-1 and Top-5 accuracy. We noticed that the Top-5 accuracy of every group of conditions are relative high, because the number of categories in ARID is too small, all the model can achieve best results on the benchmark dataset. Delta sampling shows a better performance compared with original sampling method as it samples frames in the interval of action happening in a video, reducing information loss. And we enlarge the sequences of input for the same reason.

\begin{table}[ht]
\centering
\label{tabelsample}
\begin{tabular}{cccccc}
\toprule
Train & Test & L2-norm & Top-1 & Top-5\\
\midrule
Center & Maxcenter & False & 90.67$\%$ & 98.89$\%$\\
Maxcenter & Maxcenter & False & 93.33$\%$ & 98.44$\%$\\
Scaling & Maxcenter & False & 92$\%$ & 98.89$\%$\\
Center & Center & True & 87.33$\%$ & 99.78$\%$\\
Maxcenter & Center & True & 85.78$\%$ & 99.56$\%$\\
Scaling & Center & True & 84.89$\%$ & 99.77$\%$\\
Center & Center & False & 84$\%$ & 99.78$\%$\\
Maxcenter & Center & False & 87.89$\%$ & 99.78$\%$\\
Scaling & Center & False & 85.33$\%$ & 99.78$\%$\\
Center & Maxcenter & True & 85.78$\%$ & 99.11$\%$\\
Maxcenter & Maxcenter & True & 86$\%$ & 99.56$\%$\\
Scaling & Maxcenter & True & 88$\%$ & 99.11$\%$\\
\bottomrule
\\
\end{tabular}
\caption{Top-1 and Top-5 accuracy of different resizing method in training and testing.}
\end{table}

Then we tested whether different method of resizing image will make some distribution to classification accuracy. The result is shown in Tab 2. All the Top-5 accuracy are nearly $100\%$ so this can not be a suitable for judging whether this method is good or not. When we chose maxcenter in training and testing period, it provides a highest $93.33\%$ Top-1 accuracy. It is because when resizing, maxcenter removes some useless information in the bottom of images, only contains important information in the middle, while center method loss effective pixels when getting rid of useless pixels. 

\begin{figure}
\centering
\begin{minipage}{1.0\linewidth}
\vspace{3pt}
\centerline{\includegraphics[width=\textwidth]{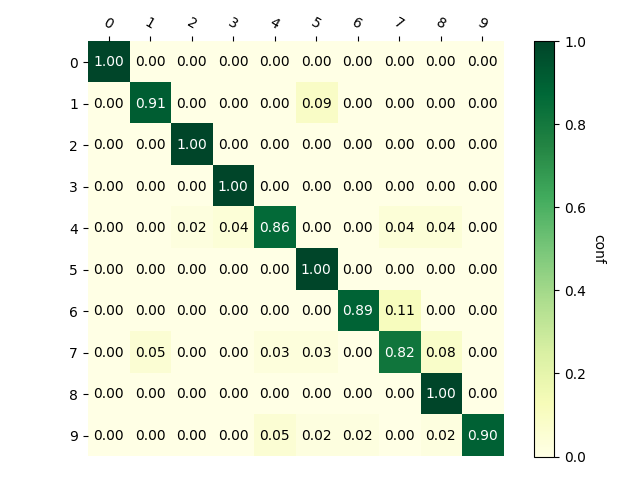}}
\end{minipage}
\caption{Recognition on different action categories.}
\label{fig}
\end{figure}
 
Finally is our designed adaptive GIC and whole model. We made this small dataset by sampling from UCF101, then changed the $\gamma$ of the original image to $\frac{1}{\gamma}$, and used $\gamma$ as the label of every image. At first we select one value $\gamma = 3$ for all samples, but the performance is not good enough. Because of this, we decided to use such a method to generate a training dataset. The Top-1 and Top-5 accuracy are shown in Tab 3, and different resizing methods were also applied. Surprising we found that in this condition maxcenter and scaling have the same Top-5 accuracy, but Top-1 in scaling in higher. This is because the information at the bottom of an image is also useful after image enhancement, but maxcenter does not take these pixels into account. Tab 4. provides the validation accuracy and testing accuracy of our model. In general, there is a softmax layer in every classification model, but in our result, the network without softmax has higher rate of recognition. We still cannot find the reason for this. Another thing need to note that there exists overfitting.One reason is that there is some noise in training samples. But the most possible reason in our mind is too many iterations in training so that noise and valueless features are considered. The recognition accuracy of our model to each category is shown in Fig. 7. It is obvious that our model has a relatively high accuracy on a given action.

\begin{table}[ht]
\centering
\label{tabelsample}
\begin{tabular}{ccc}
\toprule
Resizing & Top-1 & Top-5\\ 
\midrule
Maxcenter & 81.56$\%$ & 99.78$\%$\\
Center & 78.67$\%$ & 98.44$\%$\\
Scaling & 83.56$\%$ & 99.78\\
\bottomrule
\\
\end{tabular}
\caption{Top-1 and Top-5 accuracy on our Ind-GIC method.}
\end{table}

\section{Conclusion}
In this work we proposed a new network architecture to recognize human action in the dark environment. First, an Ind-GIC is trained to learn a suitable gamma for all training samples completing Gamma Image Correction.
Next we design a two-stream architecture, an I3D is used to extract features from optical flow and R(2+1)D used to fuse spatial and temporal features from original videos and enhanced videos. Then these two networks are stitched together to get the final feature, feeding into a BERT with multihead attention mechanism and generating a feature vector for classification. Finally, the MLP gives the recognition result. The experiment shows that our method has a relative high accuracy in the testing dataset.  

\vspace{10pt}
\begin{table}[ht]
\centering
\begin{tabular}{cccccc}
\toprule
  & No softmax & Softmax\\
\midrule
Validation & 96.88$\%$ & -\\
Testing & 93.33$\%$ & 92.89$\%$\\
\bottomrule
\\
\end{tabular}
\caption{Accuracy on out Two-stream input network.}
\end{table}

\end{document}